\documentclass[a4paper,conference]{IEEEtran}
\ifCLASSINFOpdf
\else
\fi
\usepackage{graphicx}
\usepackage{comment}
\usepackage{amsmath,amssymb} % define this before the line numbering.
\usepackage{color}
\usepackage{footnote}
\usepackage[hyphens]{url}
\usepackage{hyperref}
\definecolor{mypurple}{RGB}{148,0,211}

\begin{document}
\title{Development of Fast Refinement Detectors on \\AI Edge Platforms}

\author{\IEEEauthorblockN{Min-Kook Choi}
\IEEEauthorblockA{hutom\\
Email: mkchoi@hutom.io}
%\and
%\IEEEauthorblockN{Jaehyung Park, Jinhee Lee, and Soon Kwon}
%\IEEEauthorblockA{DGIST\\
%Email: \{stillrunning, jhlee07, soony\}@dgist.ac.kr}
\and
\IEEEauthorblockN{Heechul Jung$^*$}
\IEEEauthorblockA{KNU\\
Email: heechul@knu.ac.kr}}


%Telephone: (800) 555--1212, Fax: (888) 555--1212}

\newcommand\blfootnote[1]{%
  \begingroup
  \renewcommand\thefootnote{}\footnote{#1}%
  \addtocounter{footnote}{-1}%
  \endgroup
}


\maketitle

\begin{abstract} 
Refinement detector (RefineDet) is a state-of-the-art model in object detection that has been developed and refined based on high-end GPU systems. In this study, we discovered that the speed of models developed in high-end GPU systems is inconsistent with that in embedded systems. In other words, the fastest model that operates on high-end GPU systems may not be the fastest model on embedded boards. To determine the reason for this phenomenon, we performed several experiments on RefineDet using various backbone architectures on three different platforms: NVIDIA Titan XP GPU system, Drive PX2 board, and Jetson Xavier board. Finally, we achieved real-time performances (approximately 20 fps) based on the experiments on AI edge platforms such as NVIDIA Drive PX2 and Jetson Xavier boards. We believe that our current study would serve as a good reference for developers who wish to apply object detection algorithms to AI edge computing hardware. The complete code and models are publicly available on \href{https://github.com/mkchoi-0323/modified_refinedet}{the web (link)}. \blfootnote{$^*$corresponding author.}%\footnote{\href{https://github.com/mkchoi-0323/modified_refinedet}{link}}.
\end{abstract}

% \fi
\IEEEpeerreviewmaketitle

\section{Introduction}
%In recent years, object detection has improved considerably owing to the emergence of object detection networks that utilize the structure of convolutional neural networks (CNNs) \cite{Liu18}. Therefore, CNN-based object detectors are strong candidates for practical applications, such as video surveillance \cite{Jung17}, autonomous navigation \cite{Wu17}, machine vision \cite{Yao18}, and medical imaging \cite{Jin18}. Several industries have implemented these technological advancements in conjunction with industrial applications. However, various practical limitations in real-time scenarios still exist in terms of inference speed (processing time) when using object detection networks, especially in embedded platforms. 

In recent years, the performance of object detection has dramatically improved due to the emergence of object detection networks that utilize the structure of CNNs \cite{Liu18}. Owing to this improvement, CNN-based object detectors have potential practical applications such as video surveillance \cite{Jung17}, autonomous navigation \cite{Wu17}, machine vision \cite{Yao18}, and medical imaging \cite{Jin18}. Several industries have been making efforts to implement these technological advancements in conjunction with industrial applications.

\begin{figure}[t!]
  \centering
  \includegraphics[width=0.8\linewidth]{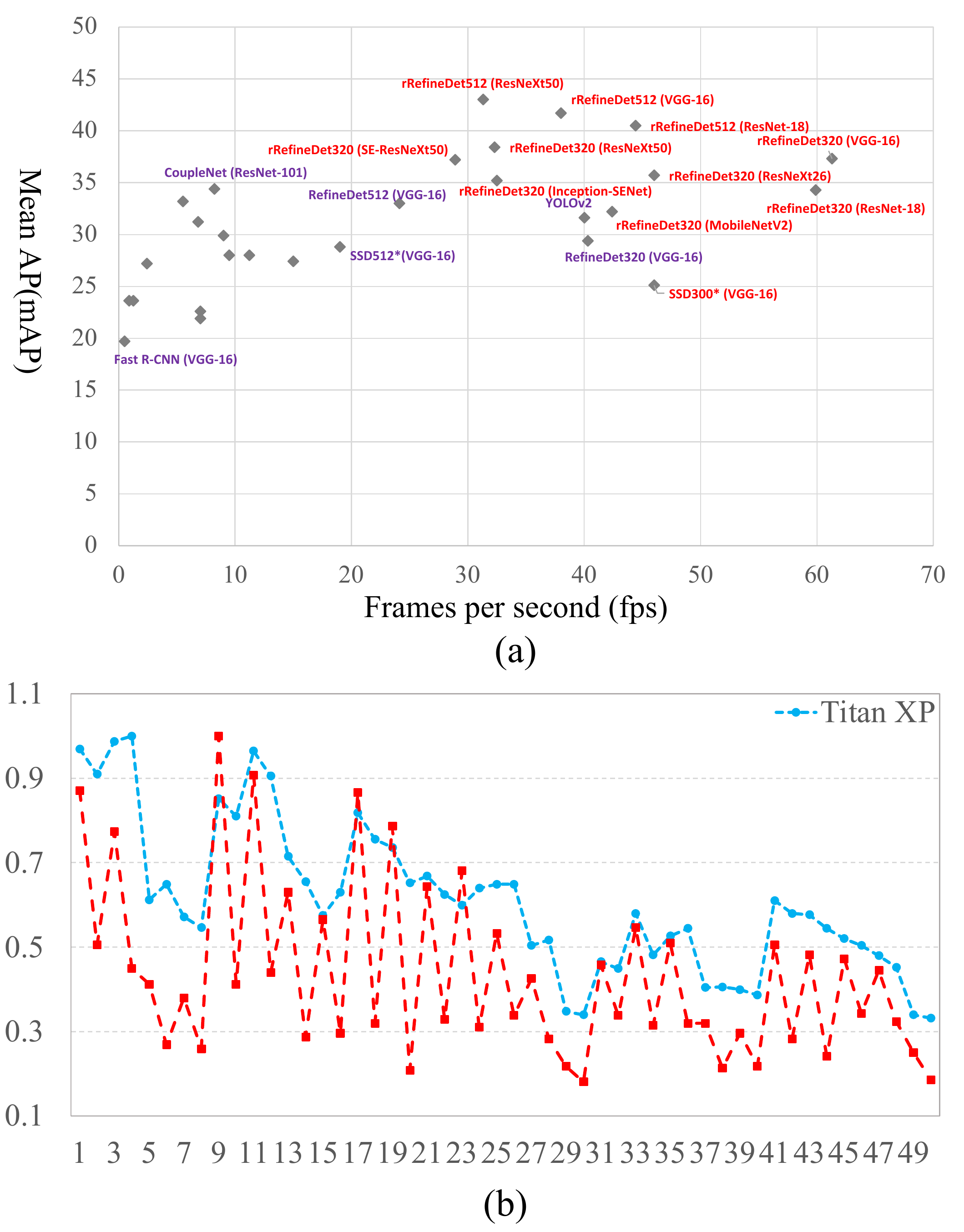}
  \vspace{-2mm}
  \caption{\textbf{Experiment results on MS-COCO dataset with Refinement Detectors.} (a) Speed (fps) versus accuracy (mAP) on MS-COCO. Out models (\textcolor{red}{red color}) have a balanced speed and accuracy compared to the existing real-time object detection network-oriented models (\textcolor{mypurple}{purple color}). Both the speed and accuracy of the proposed model were measured on the NVIDIA Titan XP. Details of the performance measurements are described in Section 4. (b) Speed (fps) variation based on Titan XP and Drive PX2 systems with the different experimental settings. X-axis represents the experiment number as shown in Table 1, and Y-axis represents speed (fps) normalized by the largest fps among 50 experiments for each system. Each experiment showed that the variation in inference speed on the edge board was much more sensitive to the training setup and architecture choices.}
\vspace{-5mm}
\end{figure}

Object detection methods using the CNN structure are broadly classified into two types: The first is a learning method by classifying class information, which locates and classifies objects in an image using a one-stage training method, by including them into one network stream and dividing the encoded features into different depth dimensions. Representative models of this one-stage network are YOLO \cite{Redmon17}, single shot multibox detector (SSD) \cite{Liu17}, SqueezeDet \cite{Wu17}, and RetinaNet \cite{Lin17}. The advantage of these networks is that the encoder of the head part leading to the network output is constructed relatively intuitively, and the inference speed is high, while the accuracy of the bounding box regression, which specifies the exact position of the object, is generally lowered.

The second strategy is a two-stage learning method in which a network that classifies approximate positions and classes of objects is searched to obtain a more precise object location using a structure in which an independent encoder within the head is separated. These two-stage net-works have a region proposal network (RPN) that can more accurately determine the existence of objects in the head of the network. R-CNN \cite{Girshick14} is the first model using this two-stage learning approach and Faster R-CNN \cite{Ren15}, R-FCN \cite{Dai16}, Deformable ConvNets \cite{Dai17}, and Mask-RCNN \cite{He17} are some of the advanced versions. The advantage of the two-stage model is greater accuracy of the bounding box regression compared with the one-stage model. However, owing to its relatively complicated head structure, the inference and training time is longer than the one-stage model.

Therefore, models with modified overall detection network structures have been proposed to compensate for the disadvantages of the one-stage and the two-stage models \cite{Li17,Zhang18}. \cite{Li17} proposed a lightweight head structure to achieve an inference speed close to that of the one-stage models. Conversely, \cite{Zhang18} suggested a model that uses the advantage of the two-stage model while maintaining the overall architecture of the one-stage model. In recent years, due to the increasing research efforts, the performance of object detection using CNN has improved dramatically in terms of accuracy and speed; however, the proposed algorithms still have various practical limitations in real-time applications in terms of accuracy and inference speed, especially in AI edge devices using embedded platforms.

Herein, we present several variations of RefineDet (refinement detector) \cite{Zhang18}, which is one of state-of-the-art object detection networks. To demonstrate how their performances vary according to their parameter settings and backbone architectures, we compare the inference performances between \textit{on high-end GPU} and \textit{on embedded platforms}. Unlike the inference on high-end GPU systems, obtaining the optimal parameters and backbone architectures is considerably important in embedded systems. Figure 1 shows the performance variation according to parameter settings and backbone architectures. Particularly, models tested on Drive PX2 exhibit a higher speed variance than high-end GPU systems, implying that hyper-parameters must be carefully selected when embedding or deploying models using the RefineDet architecture into embedded platforms. The fastest model in high-end GPU systems is not the fastest in embedded platforms.

Meanwhile, to achieve real-time inference speed with limited resources, we examined the limitations of the existing networks and applied various lightweight backbone structures to the head structure of RefineDet \cite{Zhang18} by constructing optimal intermediate layers for an effective training. The utilized backbone CNNs include VGG \cite{Liu16} and shallow ResNet \cite{He16}, which are the most widely used neural networks in image recognition. ResNeXt \cite{Xie17}, Xception \cite{Chollet17}, and MobileNet \cite{Howard17,Sandler18}, which are known for their high computation efficiency compared with layers of the same depth, were used; SENet \cite{Hu18} was applied using the “reweighting by local encoding” structure to generate punchy convolutional feature maps. To confirm the effect of feature generation on the real-time performance of the head structure, extensive comparative experiments were performed by applying head structures with a reduced channel depth for the head structure of RefineDet. Finally, we verified and confirmed the real-time performance on the NVIDIA Drive PX2 and the Jetson Xavier edge platforms using the ResNet18-based RefineDet model of reduced number of parameters and specific NMS parameters (see Table 1).

The major technical contributions of this study can be summarized as follows: 1) We present real-time detection models for AI edge platforms using a modified head structure of RefineDet. 2) The object detection network using the latest lightweight backbone with connections of the RefineDet head structure is extensively compared and analyzed using the MS-COCO 2017 detection dataset \cite{Lin14}. 3) Performance tests and model analysis were performed on NVDIA Drive PX2 and Jetson Xavier to achieve a balanced performance in terms of accuracy and speed for real-time detection on edge platforms. In addition, we introduced some issues related to the object detection models on the embedded GPU and summarized the concerns that must be addressed for achieving high real-time performance. We hope that this study will be useful for the development of real-time applications using diverse CNN-based architectures on AI edge platforms.

%%------------------------------------------------------------------------------------------------------------------------------------------------------------------------

\section{Baseline Architecture - RefineDet}
We used the basic structure of a single-shot refinement object detector (RefineDet) \cite{Zhang18} as the baseline architecture, which is derived from a one-stage learning structure of a single-shot multibox detector (SSD) \cite{Liu17}. The architectural weakness of the localization regressor of the one-stage detector was compensated using an anchor refinement module (ARM) branch, which functions similarly as the region proposal network in a two-stage detector. RefineDet minimizes the combined objective with two modules from the modified head structure of the former SSD architecture \cite{Zhang18}.%:
%\vspace{-1mm}
%\begin{equation}
%  \begin{aligned}
%  L(p_{i}, x_{i}, c_{i}, t_{i})=\\\frac{1}{N_{b}}\Big(\sum_{i=0}(L_b(p_{i},[l_{i}\geq1]))+\sum_{i=0}[l_{i}\geq1]L_{r}(x_{i},g_{i})\Big)\\+\frac{1}{N_{m}}\Big(\sum_{i=0}(L_m(c_{i},l_{i}))+\sum_{i=0}[l_{i}\geq1]L_{r}(t_{i},g_{i})\Big),
% \end{aligned}
%\end{equation}
%where the objective's index $i$ denotes the index of the anchor in the input mini-batch; objective input $a$ is the inferenced output for the binary classifier to determine the objectness; $x$ is the inferenced output of the regressor for the prediction of the bounding box location; $l$ denotes the class vector of the ground truth label to obtain loss information, and $g$ denotes the ground truth size and location. Furthermore, another objective input $c$ is the predicted confidence value for the multi-class inferenced output of the object on the feature map; $t$ is the regressor output for the location and size of the multi-class object; $N_{b}$ and $N_{m}$ denote the number of positive anchors entering each loss term; $L_{b}$ is defined as a cross-entropy log loss for binary classification; and $L_{m}$ is defined as a softmax loss for multi-class classification. The bracket indicator function $[l_{i}\geq1]$ denotes the condition when the positive anchor is true.

The connection of the head structure of RefineDet to the backbone CNNs is divided into two branches. Among them, the ARM module is a type of region proposal classifier that learns to minimize the binary classification loss of objectness and supports multiclass inference with loss generation by the backpropagation of object existence and location information. The other branch comprises an object detection module (ODM) that is used to deduce the predicted confidence and localization bounding box information for multiclass objects. The entire network is trained such that the balanced loss for both branches is minimized. To utilize the concept of a feature pyramid through information coding between the upper and lower-layer features, \cite{Zhang18} proposed a connection structure of a transfer connection block (TCB) and intermediate layers. To verify the efficiency of the network, the most widely known structure of CNNs, i.e., VGG-16 and ResNet-101, are used as the backbone for the training and evaluation results \cite{Zhang18}.

%%------------------------------------------------------------------------------------------------------------------------------------------------------------------------

\section{Combining Light-weight Backbone CNNs}

We combined state-of-the-art convolutional blocks of modified layers and structures with the lightweight backbone of the RefineDet architecture. Furthermore, we applied feature encoding blocks according to each model to the intermediate layer in the ODM branch for an efficient loss propagation between the head structure of RefineDet and the backbones. We applied VGG-16, ResNet-18, ReNeXt-26, ResNeXt-50, SE-ResNeXt-50, Inception-SENet, MobileNetV1, MobileNetV2, and Xception to the learning architecture. Furthermore, we used the feature encoding block of each backbone model as the intermediate layers. All backbone architectures were initialized through pre-training using ImageNet data \cite{Russakovsky15}. To evaluate the tradeoff between inference accuracy and speed enhancement of the detection networks, the capability of the feature pyramid was validated. Hence, we verified the original (256) and reduced (128) channel depths of the TCB and intermediate layers, and we tested additional convolutional layers. Herein, the RefineDet model with reduced channel depth is known as the “reduced RefineDet (rRefineDet).”
\\\\
\noindent \textbf{VGG-16 \cite{Liu16}}. Based on the VGG-16 model proposed in \cite{Liu16}, the $fc6$ and $fc7$ layers were transformed into the convolution layers, $conv\_fc6$ and $conv\_fc7$ respectively, through the subsampling parameters as shown in \cite{Zhang18}. To combine with the head structure of RefineDet, the subsequent layers including the last pooling layer of VGG-16 were removed and convolutions $conv6\_1$ and $conv6\_2$ were added to the top as additional convolution layers. As in \cite{Zhang18}, L2 normalization was used for the intermediate layers and some convolution layers.
\\\\
\noindent \textbf{ResNet-18 \cite{He16}}. The authors of \cite{Zhang18} applied the ResNet-101 architecture as the baseline backbone to improve the accuracy of RefineDet. In our study, RefineDet was learned using ResNet-18 pretrained by ImageNet for real-time inference as the backbone architecture. For the high-level feature encoding of backbone CNNs, the $res6$ block was added after the $res5$ block (similar to VGG-16), and the intermediate layer in the ODM branch used the residual encoding block with a channel depth of 256 in ResNet-18. For other parameter settings such as batch normalization (BN) and activations, learning was performed under the same conditions as those for ResNet-101 in \cite{Zhang18}.
\\\\
\noindent \textbf{ResNeXt-26 and 50 \cite{Xie17}}. ResNeXt is a CNN structure that uses group convolution to improve the efficiency of computation for aggregated residual transformations using identity mapping. According to \cite{Xie17}, its computational efficiency is higher than that of ResNet with the same depth, and it performed better for ImageNet data even though it uses only a few weight parameters. To effectively combine with the RefineDet head for ResNeXt-26, the outputs of $resx4$, $resx6$, and $resx8$ were used as the ARM and ODM outputs among eight $resx$ blocks. Furthermore, $resx9$ was used to process high-level features similar to those of ResNet. We added the features to the upper layer, and it could learn simultaneously by scratching. Regarding ResNeXt-50, $resx7$, $resx13$ and $resx16$ blocks of 16 $resx$ blocks were combined with the head of RefineDet, and $resx17$  was added to the upper layer. Regarding the intermediate layer, each output feature comprised the input for the feature pyramid encoded by one $resx$ block.
\\\\
\noindent \textbf{SE-ResNeXt-50 and Inception \cite{Hu18}.} 
SENet is a CNN configured to allow channel reweighting of the convolutional feature by applying a squeeze and excitation (SE) module to the output of the convolutional layer. The SE branch facilitates transformation in the depth dimension through channel-wise 1D encoding of the processed output feature. In our work, ResNeXt-50 and Inception towers with SE module were combined with the head of RefineDet and the performances of these models were verified. For SE-ResNeXt-50, the squeeze and excitation module was applied to all convolutional layers except the $conv1$ layer. To combine the heads, the output features of $conv3\_4$, $conv4\_6$, and $conv5\_3$ layers were used as inputs to the AMR and ODM, and $conv6$ blocks were added to the top layer. The channel depth of the intermediate layer was 256, and that of the final output depth of the $conv6$ block was 256. Inception-SENet comprised 10 inception blocks after the first convolution layer ($conv1$), max pooling, and the second convolution layer ($conv2$). Among the feature outputs obtained when Inception-SENet was used as the backbone CNN structure, $inception\_3b$, $inception\_4d$, $inception\_5b$ and RefineDet head were connected and the $inception\_6$ block was added to the top layer. The final output of the $inception\_6$ block had a channel depth of 256, and the intermediate layer 256, which was the same as that of SE-ResNeXt-50. %\cite{Hu18} reported the evaluation results of ImageNet by applying SE module to ResNet and Inception structures, which exhibited improved performance when included in various structures including the SE module. 
\\\\
\noindent \textbf{Xception \cite{Chollet17}.} Xception applies a depth-wise separable convolution to the inception tower to re-evaluate the learning efficiency using CNNs of the same structure as InceptionV3 \cite{Szegedy16}. The authors of \cite{Chollet17} proposed a network structure with a higher accuracy and better inference speed than InceptionV3, based on the fact that the “extreme” structure of Inception is almost equivalent in operation to the depth-wise separable convolution. To combine this Xception structure with the head of RefineDet, we used the outputs of $xception11$, $xception12$, and $conv4\_2$ as inputs to the ARM and ODM branches and added an additional $xception$ block ($xception13$) with a channel output depth of 256 to the top layer. Fur-thermore, we applied an $xception$ block with a channel depth of 256 in the intermediate layer.
\\\\
\noindent \textbf{MobileNetV1 \cite{Howard17} and V2 \cite{Sandler18}.}  In MobileNetV1, the cost of computation is minimized by applying depthwise separable convolutions such as in \cite{Chollet17} to reduce the operation cost from a typical convolution layer structure. In MobileNetV2, the depthwise separable convolution, which affects the computation time, is utilized as is, and the linear bottleneck structure is applied to minimize the burden on performance degradation. For MobileNetV1, we used the outputs of $conv4\_1$, $conv5\_5$, and $conv6$ to connect them to the head of RefineDet and added a convolution block with a channel depth of 512 to the top layer. Considering that the number of parameters of the backbone CNN structure is relatively smaller than that of the other architectures, we attempted to maintain the depth of the intermediate layers at the same feature volume as that of a basic convolutional block. Depth-wise separable convolution was applied to the top convolution block and intermediate layer. Regarding MobileNetV2, the outputs of $conv3\_2$, $conv4\_7$ and $conv6\_4$ were used as the feature outputs for the coupling with head, and the $conv7$ block was added at the top. The output channel depth of $conv7$ was 96 after narrowing down and the intermediate layer was of the same size. 

\begin{figure}[t!]
  \centering
  \includegraphics[scale=0.3]{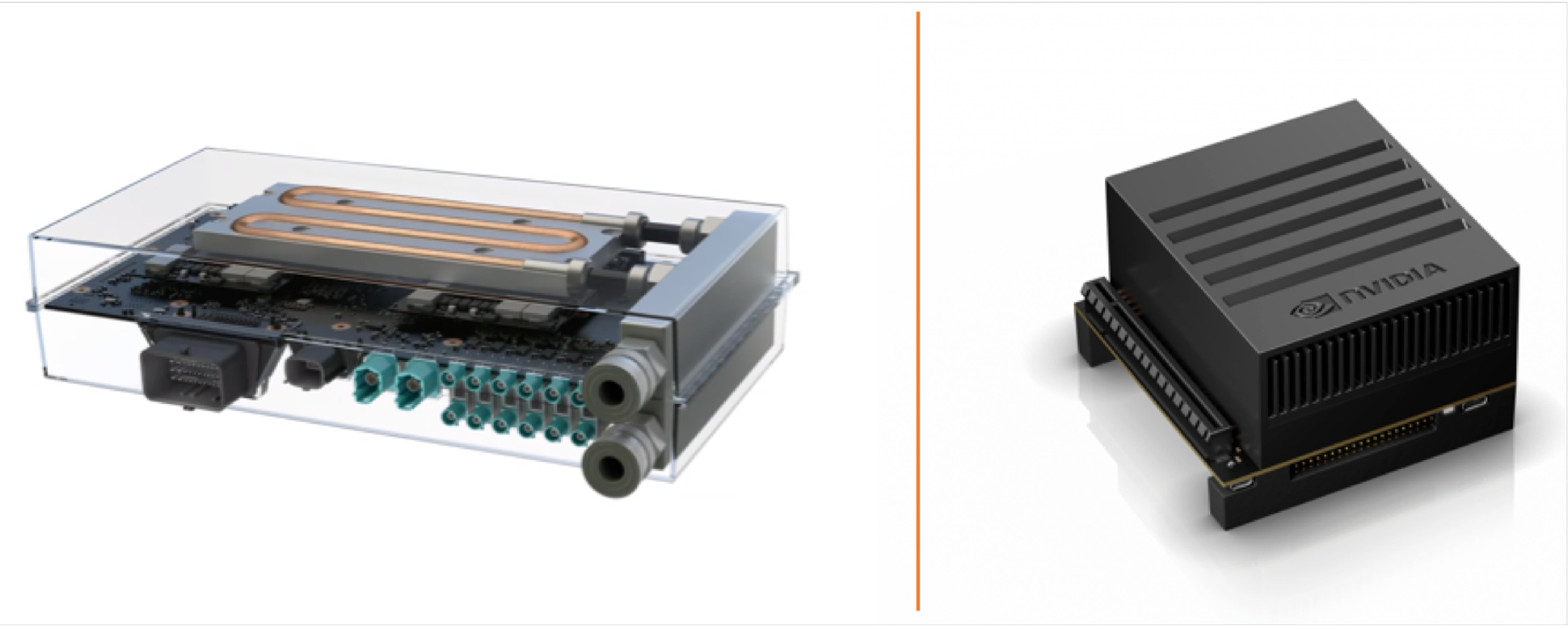}
  \caption{\textbf{NVIDIA Drive PX2 (left) and Jetson Xavier (right).} Both AI edge devices support embedded GPU based parallel computing with Pascal (Drive PX2) and Volta (Jetson Xavier) architectures.}
  \vspace{-5mm}
\end{figure}

%%------------------------------------------------------------------------------------------------------------------------------------------------------------------------
\section{Evaluation Environments}

\noindent \textbf{Dataset Preparation.} We used the MS-COCO 2017 object detection dataset \cite{Lin14} to evaluate the performances of various RefineDet networks. The dataset comprises 80 object classes, approximately 120,000 training images ($train17$), 5,000 validation images ($val17$), 80,000 test images ($test17$), and 120,000 unlabeled images ($unlabeled17$). For the evaluation of the proposed models, all models were trained with $train17$. Moreover, $val17$ and $test\textnormal{-}dev17$ were used as test sets, separately. For the quantitative evaluations of $val17$ and $test\textnormal{-}dev17$, the mean average precision (mAP) value was used as the evaluation criterion. The final mAP was calculated as the mean value obtained from a range of intersection over union (IoU) with $[0.5: 0.05: 0.95]$.
\\\\
\noindent \textbf{Plaforms.} %Also, we evaluated the inference speed of the models in the embedded environments of NVIDIA Drive PX2 and Jetson Xavier.
We used Titan XP, Drive PX2, and Jetson Xavier as our testing platform. Both platforms were developed for AI edge computing, and specifically for DrivePX2, they are designed for application to autonomous vehicles. We performed training on the Titan XP, Drive PX2, and Jetson Xavier systems. Drive PX2 is based on the Tegra X2 SoC board and contains 12 CPU cores: eight of A57 and four of Denver. The Pascal architecture of the GPU processor is based on the 16FinFET process and supports UART, CAN, LIN, FlexRay, USB, and 1 or 10 Gbit Ethernet communication (see Figure 2). Jetson Xavier is a recently released embedded architecture based on a 512-Core Volta GPU with tensor cores and an 8-core ARM CPU. This architecture has 16 GB 256-bit of LPDDR4x memory and 32 GB of eMMC 5.1 flash storage and supports (2x) NVDLA DL accelerator engines.
\\\\
\noindent \textbf{Training.} For training each model quickly, we used the pretrained weights as the initialization from ImageNet. The top layers of the convolution blocks, intermediate layers, and TCB of the RefineDet structure for the feature pyramid were initialized to a random Gaussian distribution with $\sigma = 0.01$. For a fair performance evaluation of the proposed models, the same parameters related to the detection model (anchor size, IoU with ground truth box, types of training data augmentation, etc.) were used except for the modification of the backbone and head structures. In addition, to confirm the role of the last convolutional block for high-level feature processing and the TBC for the feature pyramid, the channel depth of each feature map was divided into two cases of 128 and 256. The total learning duration was 120 epochs, and stochastic gradient descent was used as an optimizer. The base learning rate was started at 0.001, and a drop rate of 0.1 to 84 and 108 epochs was applied. The weight decay was 0.0005 and the momentum was 0.9. All the learning was performed in the Caffe environment using the Python interface. The depthwise separable convolution utilized the implementation code in the Caffe environment available to the public and the original code for the customizing layer of the SE module. In order to evaluate the performance of the model under various conditions, we trained different models according to the feature depth and input image size for TCB and top layers. 
\\\\
\noindent \textbf{Testing.} We performed testing process using $val17$ and $test\textnormal{-}dev17$ datasets for each model. Additionally, we used several non-maxima suppression (NMS) parameters to demonstrate the effectiveness of the NMS according to the system. rRefineDet is a training model that focuses on the inference speed and limits the size of the feature depths of the TCB, intermediate layer, and top layer. As a reference, the evaluation results of MS-COCO $test\textnormal{-}dev14$ data are included to show the results of original four RefineDet models reported in \cite{Zhang18}. Unlike our experiments, the experiments reported in the original RefineDet paper were performed on a Titan X GPU. Therefore, a lower fps was indicated compared with our experiments even when the same model and parameter settings were used. The average for every inference speed was obtained based on 5,000 input images ($10$ for warm-up).

\section{Results and Discussions}
%\noindent \textbf{Titan XP VS NVIDIA Drive PX2.} %Also, we evaluated the inference speed of the models in the embedded environments of NVIDIA Drive PX2 and Jetson Xavier.

\begin{table*}
  \caption{Performance change in different output feature depths from top layers of the high-level convolutional block, TBC and intermediate layer and difference in inference speed according to the NMS parameter on each platform. rRefineDet is a model that minimizes the feature output depth of the top convolutional blocks, TBC, and intermediate layers. NMS parameters in the NMS column include the maximum number of candidate bounding boxes for the NMS input, the maximum number of predicted bounding boxes according to the NMS output, and the confidence threshold for the final output. Every model was trained using $train17$ and tested on the Drive PX2 embedded platform. $td$ means $test\textnormal{-}dev$. The performance results of the first four models were extracted from the original RefineDet paper \cite{Zhang18}. Red, blue, and green colors represent the 1st, 2nd, and 3rd fastest models among 50 experiments, respectively.}
  \label{tab:freq}
  \resizebox{\textwidth}{!}{
  \begin{tabular}{c|c|c|c|c|c|c|c}
  %  \toprule
    Exp. No. & Model & Backbone & NMS parameters & mAP ($val17$) & mAP ($td17$) & fps (Titan XP) & fps (Drive PX2) \\ \hline\hline
 %   \midrule
\cite{Zhang18} & RefineDet320  & VGG-16 & (1000, 500, 0.01) & - & 29.4 ($td14$) & 40.3 (Titan X) & -\\
\cite{Zhang18} & RefineDet320  & VGG-16 & (1000, 500, 0.01) & - & 33 & 24.1 (Titan X) & -\\
\cite{Zhang18} & RefineDet320  & ResNet-101 & (1000, 500, 0.01) & - & 32 & 9.3 (Titan X) & -\\
\cite{Zhang18} & RefineDet320  & ResNet-101 & (1000, 500, 0.01) & - & 36.4 & 5.2 (Titan X) & -\\ \hline
    1 & rRefineDet320 & VGG-16 & (400, 200, 0.1) & 31.8 & 36.1& \textcolor{green}{\textbf{60.2}} &  \textcolor{green}{\textbf{18.8}} \\ 
    2 & rRefineDet320 & VGG-16 & (1000, 500, 0.01) & 32.7 & 37.2 & 56.5 & 10.9 \\ 
    3 & RefineDet320 & VGG-16 & (400, 200, 0.1) & 32.2  & 36.4 & \textcolor{blue}{\textbf{61.3}} & 16.7 \\
    4 & RefineDet320 & VGG-16 & (1000, 500, 0.01) & 33.7 & 37.3 & \textcolor{red}{\textbf{62.1}} & 9.7 \\
    5 & rRefineDet512 & VGG-16 & (400, 200, 0.1) & 34.4 & 40.7 & 38.0 & 8.9 \\
    6 & rRefineDet512 & VGG-16 & (1000, 500, 0.01) & 35.5 & 41.7 & 40.3 & 5.8 \\
    7 & RefineDet512 & VGG-16 & (400, 200, 0.1) & 35.4 & 40.9 & 35.5 & 8.2 \\
    8 & RefineDet512 & VGG-16 & (1000, 500, 0.01) & 36.4 & 42.0 & 33.9 & 5.6 \\ \hline
    9 & rRefineDet320 & ResNet-18 & (400, 200, 0.1) & 31.0 & 33.5 & 52.9 & \textcolor{red}{\textbf{21.6}} \\
    10 & rRefineDet320 & ResNet-18 & (1000, 500, 0.01) & 31.7 & 34.3 & 50.3 & 8.9 \\
    11 & RefineDet320 & ResNet-18 & (400, 200, 0.1) & 32.4 & 34.2 & 59.9 & \textcolor{blue}{\textbf{19.6}} \\ 
    12 & RefineDet320 & ResNet-18 & (1000, 500, 0.01) & 33.6 & 35.1 & 56.2 & 9.5 \\ 
    13 & rRefineDet512 & ResNet-18 & (400, 200, 0.1) & 35.8 & 40.2 & 44.4 & 13.6 \\ 
    14 & rRefineDet512 & ResNet-18 & (1000, 500, 0.01) & 36.2 & 40.5 & 40.7 & 6.2 \\ 
    15 & RefineDet512 & ResNet-18 & (400, 200, 0.1) & 36.6 & 40.3 & 35.7 & 12.2 \\ 
    16 & RefineDet512 & ResNet-18 & (1000, 500, 0.01) & 37.6 & 41.5 & 39.1 & 6.4 \\ \hline
    17 & rRefineDet320 & MobileNetV1 & (400, 200, 0.1) & 26.2 & 30.0 & 50.8 & 18.7 \\
    18 & rRefineDet320 & MobileNetV1 & (1000, 500, 0.01) & 27.2 & 30.8 & 46.9  & 6.9 \\
    19 & RefineDet320 & MobileNetV1 & (400, 200, 0.1) & 28.2 & 31.3 & 45.7 & 17.0 \\ 
    20 & RefineDet320 & MobileNetV1 & (1000, 500, 0.01) & 30.0 & 32.0 & 40.5 & 4.5 \\ \hline 
    21 & rRefineDet320 & MobileNetV2 & (400, 200, 0.1) & 26.7 & 30.8 & 41.5 & 13.9 \\
    22 & rRefineDet320 & MobileNetV2 & (1000, 500, 0.01) & 27.5 & 31.1 & 38.8 & 7.1 \\
    23 & RefineDet320 & MobileNetV2  & (400, 200, 0.1) & 28.5 & 32.2& 37.2 & 14.7 \\
    24 & RefineDet320 & MobileNetV2  & (1000, 500, 0.01) & 29.2 & 33.0 & 39.7 & 6.7 \\ \hline
    25 & rRefineDet320 & Inception-SENet & (400, 200, 0.1) & 33.1 & 32.5 & 40.3 & 11.5 \\ 
    26 & rRefineDet320 & Inception-SENet & (1000, 500, 0.01) & 33.2 & 28.3 & 40.3 & 7.3 \\ 
    27 & RefineDet320 & Inception-SENet & (400, 200, 0.1) & 34.2 & 35.0 & 31.3 & 9.2 \\
    28 & RefineDet320 & Inception-SENet & (1000, 500, 0.01) & 35.2 & 35.8 & 32.1 & 6.1 \\
    29 & rRefineDet512 & Inception-SENet & (400, 200, 0.1) & 37.4 & 41.8 & 21.6 & 4.7 \\
    30 & rRefineDet512 & Inception-SENet & (1000, 500, 0.01) & 37.7 & 42.1 & 21.1 & 3.9 \\ 
    31 & rRefineDet320 & SEResNeXt-50 & (400, 200, 0.1) & 35.2 & 37.2 & 28.9 & 9.9 \\ 
    32 & rRefineDet320 & SEResNeXt-50 & (1000, 500, 0.01) & 36.1 & 38.2 & 27.9 & 7.3 \\ \hline     
    33 & rRefineDet320 & ResNeXt-26 & (400, 200, 0.1) & 30.5 & 34.9 & 36.0 & 11.8 \\
    34 & rRefineDet320 & ResNeXt-26 & (1000, 500, 0.01) & 31.3 & 35.7 & 29.9 & 6.8 \\
    35 & RefineDet320 & ResNeXt-26 & (400, 200, 0.1) & 32.1 & 35.7 & 32.7 & 11.0 \\ 
    36 & RefineDet320 & ResNeXt-26 & (1000, 500, 0.01) & 32.8 & 36.6 & 33.8 & 6.9 \\ 
    37 & rRefineDet512 & ResNeXt-26 & (400, 200, 0.1) & 34.4 & 39.5 & 25.1 & 6.9 \\
    38 & rRefineDet512 & ResNeXt-26 & (1000, 500, 0.01) & 35.4 & 40.5 & 25.2 & 4.6 \\
    39 & RefineDet512 & ResNeXt-26 & (400, 200, 0.1) & 35.2 & 40.3 & 24.8 & 6.4 \\
    40 & RefineDet512 & ResNeXt-26 & (1000, 500, 0.01) & 36.2 & 41.3 & 24.0 & 4.7 \\ \hline
    41 & rRefineDet320 & Xception & (400, 200, 0.1) & 34.6 & 37.2 & 37.9 & 10.9 \\
    42 & rRefineDet320 & Xception & (1000, 500, 0.01) & 35.0 & 38.0 & 36.0 & 6.1 \\
    43 & RefineDet320 & Xception & (400, 200, 0.1) & 34.9 & 37.8 & 35.8 & 10.4 \\
    44 & RefineDet320 & Xception & (1000, 500, 0.01) & 35.7 & 38.9 & 33.8 & 5.2 \\ \hline
    45 & rRefineDet320 & ResNeXt-50 & (400, 200, 0.1) & 35.6 & 37.6 & 32.3 & 10.2 \\
    46 & rRefineDet320 & ResNeXt-50 & (1000, 500, 0.01) & 36.1 & 38.4 & 31.3 & 7.4 \\
    47 & RefineDet320 & ResNeXt-50 & (400, 200, 0.1) & 36.7 & 38.0 & 29.8 & 9.6 \\
    48 & RefineDet320 & ResNeXt-50 & (1000, 500, 0.01) & 37.6 & 38.4 & 28.1 & 7.0 \\
    49 & rRefineDet512 & ResNeXt-50 & (400, 200, 0.1) & 36.9 & 42.5 & 21.1 & 5.4 \\
    50 & rRefineDet512 & ResNeXt-50 & (1000, 500, 0.01) & 37.8 & 43.5 & 20.6 & 4.0 \\   
%  \bottomrule
\end{tabular}
}
\end{table*}

\noindent \textbf{Changing Backbone Architecture.} 
As shown in Table 1, changing the structure of the feature connection blocks between the head and backbone affected the performance. Models with VGG-16, ResNet-18 as the backbone structure demonstrated superior inference speed compared with the ResNeXt-26, MobileNetV1, and V2 models. Particularly, VGG-16 was the fastest architecture on Titan XP, but ResNet-18 was the fastest on Drive PX2. By combining the head structure of the existing object detection networks and CNN models, a balanced performance can be achieved, which is applicable to various fields. Moreover, the quantitative performance can be enhanced by setting a different input size for the image and applying the deformable operation to a specific convolutional layer according to the available computing resources. In addition, it is clear that using the improved convolutional block such as feature renormalization of the SE module and the inverted residual structure of MobileNetV2 in a specific head and backbone structure helps to prevent speed degradation and improve accuracy. 
\\\\
\noindent \textbf{Effectiveness of NMS Parameters.} As shown in Table 1, a slight adjustment of the NMS parameter resulted in a considerable improvement in the inference speed at the expense of a low accuracy. Furthermore, unexpected computational bottlenecks appeared when applying a state-of-the-art algorithm based on CNNs in a Drive PX platform. Layer-wise inference testing was performed on all layers of the RefineDet architecture to analyze the cause. We discovered that the computation times of all layers operating on the embedded GPU increased linearly compared with the number of high-end GPU cores, but the post-processing for the bounding box occupied most of the bottlenecks. Because the model is designed to process bounding box filtering related to the NMS operation of the post-processing layer, the CPU operations on a certain computing platform causes a severe performance degradation compared with high-end GPU systems rich in CPU computing resources. Table 1 shows that performance degradation owing to the CPU operation bottleneck can cause a significant performance degradation not only in embedded environments, but also in general computing resources. Furthermore, Table 1 shows the comparison results of the inference speed on various platforms according to the model structure and NMS parameters. To reduce the burden of the CPU operations, the input bounding box for NMS is set to a maximum of 400, and the confidence threshold for the output to the NMS is increased to 0.1. Although the accuracies are not significantly different, the performance gain in the Drive PX2 environment is extremely high owing to the adjustment of the NMS parameter, which is not significant in the high-end GPU system environment. The tradeoff between speed and performance can be improved significantly by identifying the location of the computational burden in the architecture and adjusting the hyperparameter for the related operation.
\\\\
\noindent \textbf{NVIDIA Drive PX2 vs. Jetson Xavier.}
As our experiments were primarily performed on Drive PX2, we also validated how RefineDet operates on different edge platforms. First, we select two architectures of VGG-16 and ResNet-18 for the comparison, because those backbones are the fastest models on Titan XP and Drive PX2, respectively. Interestingly, ResNet-18, which is the fastest model for Drive PX2, is also the fastest model for Jetson Xavier. This confirms that the fastest network in a high-end system can produce different results on an embedded board. Meanwhile, although the hardware specifications of Drive PX2's GPU are better than those of Jetson Xavier’s GPU, Jetson Xavier shows similar or better performances in some models. We conjecture that this is because Jetson Xavier contains a deep learning accelerator. Finally, the speed variation is more stable in Jetson Xavier, which appears to have a more stable hardware architecture for a balanced computation resource between CPU and GPU cores.
\\\\
\noindent \textbf{mAP with SOTA detectors.} For reference to detection accuracy, Table 3 shows mAPs with state-of-the-art models. We note that depending on the training setup of each model, different accuracy can be achieved despite the same model.

\begin{table}[h]
\vspace{-3mm}
  \caption{Inference speed of NVIDIA Drive PX2 vs. Jetson Xavier (Jetson-X). Index of experiments is the same as that in Table 1.}\vspace{-2mm}
  \label{tab:freq}
  \resizebox{\linewidth}{!}{
  \begin{tabular}{c|c|c|c|c}
%    \toprule
\hline
   Exp. No. & Model & Backbone & fps (Drive PX2) & fps (Jetson-X)\\ \hline
 %   \midrule
    1 & rRefineDet320 & VGG-16 & \textcolor{green}{\textbf{18.8}} & 13.6\\ 
    2 & rRefineDet320 & VGG-16 & 10.9 & 10.7\\ 
    3 & RefineDet320 & VGG-16 & 16.7 & 12.5\\
    4 & RefineDet320 & VGG-16 & 9.7 & 11.4\\
    5 & rRefineDet512 & VGG-16 & 8.9 & 6.6\\
    6 & rRefineDet512 & VGG-16 & 5.8 & 6.0\\
    7 & RefineDet512 & VGG-16 & 8.2 & 7.6\\
    8 & RefineDet512 & VGG-16 & 5.6 & 6.9\\ \hline
    9 & rRefineDet320 & ResNet-18 & \textcolor{red}{\textbf{21.6}} & \textcolor{red}{\textbf{22.1}}\\
    10 & rRefineDet320 & ResNet-18 & 8.9 & \textcolor{blue}{\textbf{18.5}}\\
    11 & RefineDet320 & ResNet-18 & \textcolor{blue}{\textbf{19.6}} &  \textcolor{green}{\textbf{17.0}}\\ 
    12 & RefineDet320 & ResNet-18 & 9.5 & 16.6\\ 
    13 & rRefineDet512 & ResNet-18 & 13.6 & 11.9\\ 
    14 & rRefineDet512 & ResNet-18 &6.2 & 11.5\\ 
    15 & RefineDet512 & ResNet-18 & 12.2 & 11.2\\ 
    16 & RefineDet512 & ResNet-18 & 6.4 & 11.1\\ %\hline
%  \bottomrule
\hline
\end{tabular}
}
\vspace{-3mm}
\end{table}
\begin{table}
  \caption{Accuracy of state-of-the-art detectors using the MS-COCO dataset. The order is sorted by accuracy and speed. $td$ means $test\textnormal{-}dev$. We note that even on the same architecture, performance differences may occur due to different settings.}
%  \vspace{-2mm}
  \label{tab:freq}
  \resizebox{.49\textwidth}{!}{
  \begin{tabular}{c|c|c|c}
    Model&Backbone& training & mAP (data) \\ \hline
    Fast R-CNN \cite{Girshick15} & VGG-16 & $train14$ & 19.7 ($td14$) \\ 
    Faster R-CNN \cite{Ren15} & VGG-16 & $trainval14$ & 21.9 ($td14$) \\ 
    R-FCN \cite{Dai16} & ResNet-101 & $trainval14$ & 29.9 ($td14$) \\ 
    Def. R-FCN \cite{Dai17} & ResNet-101 & $trainval14$ & 34.5 ($td14$) \\ 
    Def. R-FCN \cite{Dai17} & Inception-ResNet & $trainval14$ & 37.5 ($td14$) \\
    umd\_det \cite{Bodla17} & ResNet-101 & $trainval14$ & 40.8 ($td14$) \\
    G-RMI \cite{Huang17} & Ensemble & $trainval14$ (32k) & 41.6 ($td14$) \\
    \hline
%    SSD300 \cite{Liu17} & VGG-16 & $trainval14$ (35k) & 25.1 ($td14$) \\
%    RON384++ \cite{Kong17} & VGG-16 & $trainval14$ & 27.4 ($td14$) \\
    SSD321 \cite{Liu17} & ResNet-101 & $trainval14$ (35k) & 28.0 ($td14$) \\
    DSSD321 \cite{Fu17} & ResNet-101 & $trainval14$ (35k) & 28.0 ($td14$) \\
%    SSD512 \cite{Liu17} & VGG-16 & $trainval14$ (35k) & 28.8 ($td14$) \\
    SSD513 \cite{Liu17} & ResNet-101 & $trainval14$ (35k) & 31.2 ($td14$) \\
    YOLOv2 \cite{Redmon17} & Darknet-19 & $trainval14$ (35k) & 31.6 ($td14$) \\
    RetinaNet500 \cite{Lin17} & ResNet-101 & $trainval14$ (35k) & 32.0 ($td14$) \\
    DSSD513 \cite{Fu17} & ResNet-101 & $trainval14$ (35k) & 33.2 ($td14$) \\
    RetinaNet800 \cite{Lin17} & ResNet-101-FPN & $trainval14$ (35k) & 36.4 ($td14$) \\
    \hline
    RefineDet320 \cite{Zhang18} & VGG-16 & $trainval14$ (35k) & 29.4 ($td14$) \\
    RefineDet320 \cite{Zhang18} & ResNet-101 & $trainval14$ (35k) & 32.0 ($td14$) \\
    RefineDet512 \cite{Zhang18} & VGG-16 & $trainval14$ (35k) & 33 ($td14$) \\
    RefineDet512 \cite{Zhang18} & ResNet-101 & $trainval14$ (35k) & 36.4 ($td14$) \\
    \hline
%    RefineDet320 & MobileNetV1 & $train17$ & 32.0 ($td17$) \\
    RefineDet320 & MobileNetV2 & $train17$ & 33.0 ($td17$) \\
    RefineDet320 & ResNet-18 & $train17$ & 35.1 ($td17$) \\
    RefineDet320 & Inception-SENet & $train17$ & 35.8 ($td17$) \\    
    RefineDet320 & ResNeXt-26 & $train17$ & 36.6 ($td17$) \\
    RefineDet320 & VGG-16 & $train17$ & 37.2 ($td17$) \\
    RefineDet320 & Xception & $train17$ & 38.0 ($td17$) \\
    rRefineDet320 & SEResNeXt-50  & $train17$ & 38.2 ($td17$) \\
%    RefineDet320 & ResNeXt-50 & $train17$ & 38.4 ($td17$) \\ 
    RefineDet512 & ResNeXt-26  & $train17$ & 41.3 ($td17$) \\
    RefineDet512 & ResNet-18 & $train17$ & 41.5 ($td17$) \\
    RefineDet512 & VGG-16 & $train17$ & 42.0 ($td17$) \\
    rRefineDet512 & Inception-SENet  & $train17$ & 42.1 ($td17$) \\
    rRefineDet512 & ResNeXt-50 & $train17$ & 43.5 ($td17$) \\
\end{tabular}
}
%\vspace{-5mm}
\end{table}

\section{Conclusions and Future Works}

In this study, we conducted several experiments and discovered that the choice of NMS parameters and the backbone architecture in AI edge hardware were important. Results indicated real-time performances (approximately 20 fps) on embedded platforms such as NVIDIA Drive PX2 and Jetson Xavier boards. For future studies, it is necessary to closely analyze the relationship between the characteristics of the backbone layer of our models and those of the head structure. Hence, it is necessary to perform an ablation study on each backbone structure and TBC layer or intermediate layer to analyze the importance of each relation about connecting to the head. Furthermore, we intend to apply half or mixed precision techniques such as TensorRT to obtain better optimizations on platforms with limited resources.

%\begin{figure}[t!]
% \centering
%  \includegraphics[width=\linewidth]{figs/fig4.pdf}
%  \caption{\textbf{Qualitative results on proposed models on the MS-COCO $val17$}. We denote each model as (model, backbone, size, nms parameter): from the top (RefineDet, MobileNetV2, 320, 1000), (rRefineDet, MobileNetV2, 320, 400), (rRefineDet, ResNet18, 320, 400), (RefineDet, ResNet18, 512, 1000), (rRefineDet, VGG16, 320, 400), (RefineDet, VGG16, 512, 1000), (rRefineDet, ResNeXt26, 320, 400), (RefineDet, ResNeXt26, 512, 1000), (rRefineDet, ResNeXt50, 320, 400), (rRefineDet, ResNeXt50, 512, 1000), (rRefineDet, Inception-SENet, 320, 400), and (rRefineDet, Inception-SENet, 512, 1000). Confidence threshold is set to 0.6 for better visualization. Best viewed in color.}
%\vspace{-5mm}
%\end{figure}

\end{document}